\documentclass[10pt,twocolumn,letterpaper]{article}
\usepackage{iccv}

\usepackage{times}
\usepackage{epsfig}
\usepackage{graphicx}
\usepackage{amsmath}
\usepackage{amssymb}
\usepackage{multirow}
\usepackage{makecell}
\usepackage{booktabs}
\usepackage[noadjust]{cite}

\usepackage{enumitem}
\setlist[itemize]{align=parleft,left=0pt..1em}

\usepackage{array} 
\usepackage{colortbl}%
\newcommand{\myrowcolour}{\rowcolor[gray]{0.9}}

\usepackage[pagebackref=true,breaklinks=true,colorlinks,bookmarks=false]{hyperref}

\newcommand{\figref}[1]{Fig.~\ref{#1}}
\newcommand{\equref}[1]{Eq.~\eqref{#1}}
\newcommand{\secref}[1]{Sec.~\ref{#1}}
\newcommand{\tabref}[1]{Table~\ref{#1}}

\def\etalcite#1{\etal~\cite{#1}}

\usepackage[capitalize]{cleveref}
\crefname{section}{Sec.}{Secs.}
\Crefname{section}{Section}{Sections}
\Crefname{table}{Table}{Tables}
\crefname{table}{Tab.}{Tabs.}

\newcommand{\note}[1]{\textcolor{black}{#1}}
\newcommand{\redfont}[1]{\textcolor{red}{\textbf{#1}}}
\newcommand{\bluefont}[1]{\textcolor{blue}{\textbf{#1}}}

\iccvfinalcopy %

\def\iccvPaperID{7190} %

\ificcvfinal\fi

\usepackage[accsupp]{axessibility}

\begin{document}

\title{LRRU: Long-short Range Recurrent Updating Networks for Depth Completion}

\def\SP{~~}

\author{Yufei Wang$^{1}$,
\SP
Bo Li$^{1}$,
\SP
Ge Zhang$^{1}$,
\SP
Qi Liu$^{1}$,
\SP
Tao Gao$^{2}$
\SP
Yuchao Dai$^{1}$
\\[0.1325cm]
$ ^1$Northwestern Polytechnical University and Shaanxi Key Laboratory of\\  Information Acquisition and Processing
\SP $^2$Chang'an University 
}

\maketitle
\ificcvfinal\thispagestyle{empty}\fi

\begin{abstract}
Existing deep learning-based depth completion methods generally employ massive stacked layers to predict the dense depth map from sparse input data. Although such approaches greatly advance this task, their accompanied huge computational complexity hinders their practical applications. To accomplish depth completion more efficiently, we propose a novel lightweight deep network framework, the Long-short Range Recurrent Updating (LRRU) network. Without learning complex feature representations, LRRU first roughly fills the sparse input to obtain an initial dense depth map, and then iteratively updates it through learned spatially-variant kernels. Our iterative update process is content-adaptive and highly flexible, where the kernel weights are learned by jointly considering the guidance RGB images and the depth map to be updated, and large-to-small kernel scopes are dynamically adjusted to capture long-to-short range dependencies. Our initial depth map has coarse but complete scene depth information, which helps relieve the burden of directly regressing the dense depth from sparse ones, while our proposed method can effectively refine it to an accurate depth map with less learnable parameters and inference time. Experimental results demonstrate that our proposed LRRU variants achieve state-of-the-art performance across different parameter regimes. In particular, the LRRU-Base model outperforms competing approaches on the NYUv2 dataset, and ranks 1st on the KITTI depth completion benchmark at the time of submission. 
Project page: \href{https://npucvr.github.io/LRRU/}{https://npucvr.github.io/LRRU/}.

\end{abstract}

\section{Introduction} \label{sec::introduction}

Acquiring accurate and dense scene depth plays a fundamental role in various applications, such as autonomous driving~\cite{wang2018networking} and augmented reality~\cite{NewcombeIHMKDKSHF11}.
However, existing depth sensors have inevitable limitations for both indoor and outdoor scenes~\cite{xie2022survey},
for example, the depth acquired by LiDAR is too sparse to be used directly.
Thus, depth completion, \ie, estimating the dense depth maps from sparse distance measurements, has attracted extensive research interests in industry and research communities. 

\begin{figure}[!t]
	\centering
	\includegraphics[width=0.47\textwidth]{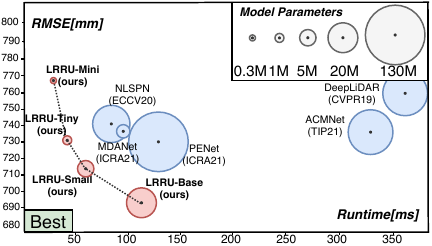}
	\caption{The performance in terms of RMSE versus runtime on the KITTI test dataset. The bubble size represents the number of parameters. LRRU performs better while maintaining efficiency.}
	\label{fig:overview}
	\vspace{-5mm}
\end{figure}

Recently, deep learning-based methods~\cite{uhrig2017sparsity, wang2022ral} have shown dominant performance for this task, which directly maps sparse depth maps to dense depth maps through massive stacked filters and layers.
Since the RGB images contain rich semantic cues that are critical for filling unknown depth, some works~\cite{ma2018sparse, MaCK19, tang2019learning, yan2021rignet} utilize the RGB information to guide depth completion.
Although many advanced networks such as ResNet~\cite{he_ResNet_CVPR_2016} and Transformer~\cite{han2022survey,rho2022guideformer, zhang2023completionformer} have been exploited, it is still difficult to directly predict an accurate and dense depth map from a sparse input depth map and the corresponding RGB image.
Specifically, existing methods employ tens of millions of learnable parameters in exchange for a desirable model capacity to learn robust features, for example, 132M parameters are contained in PENet~\cite{HuWLNFG21}.
Such large-scale networks usually require heavy computing resources, which fail to be applied in the real world, while the method performance drops significantly if the network size is simply reduced~\cite{MDANet}.
In addition, the predicted depth maps obtained by direct regression suffer from blur effect and distortion of object boundaries, which need to be further refined through extra refinement modules~\cite{cheng2019cspn}.
For example, the popular spatial propagation networks (SPNs)~\cite{cheng2019cspn,cheng2020cspn++,xu2020deformable,park2020non,HuWLNFG21,lin2022dynamic,liu2022graphcspn} update the output of the direct-regression methods by a recurrent operation.
Thus, designing an efficient depth completion architecture that performs better while maintaining efficiency is essential for further research.

Inspired by the success of the recurrent design~\cite{TeedD20, Lipson21, wang2022efficient}, we propose the Long-short Range Recurrent Updating (LRRU) network, a novel lightweight deep network framework for depth completion.
Unlike existing direct-regression methods, LRRU iteratively updates an initial depth map obtained by a non-learning approach~\cite{ku2018defense}.
The initial depth map has coarse but complete scene depth information, which can help relieve the heavy burden of directly regressing the precise dense depth from sparse input depth map.
Although existing SPNs~\cite{cheng2019cspn,cheng2020cspn++,xu2020deformable,park2020non,HuWLNFG21,lin2022dynamic,liu2022graphcspn} have shown that the depth map can be refined by learned spatially-variant kernels that model relevant neighbors and their affinities of each pixel, these methods cannot be directly used in our framework due to the following limitations: {(a) content-agnostic update unit}: the kernel parameters required for updating are predicted by the features from RGB and sparse depth, which are not adaptively adjusted to the target map (the depth map to be updated);
{(b) inflexible recurrent strategy}: the kernel scope is fixed during the update process, and multiple iterations are required to obtain long-range dependencies and satisfactory results.

To address the above issues, we propose a Target-Dependent Update (TDU) unit and a long-short range recurrent strategy, which make our iterative update process \emph{content-adaptive and highly flexible}.
\note{Our TDU predicts the sampling position of the neighbors and the weights (affinities) between them and the reference point by jointly considering the cross-guided and self-guided features.
The cross-guided features extracted from RGB images and sparse depths can guide the TDU to avoid irrelevant neighbors, while the self-guided features extracted from the depth map to be updated allow the TDU to be adaptive to the content of the target map.}
Moreover, our TDU further improves the performance by learning the residual.
In addition, we observe that when multiple TDUs of the update process employ the cross-guided features of different scales respectively, the TDU guided by smaller scale cross-guided features will adaptively learn to obtain the neighbors in a relatively large scope, and vice versa.
\note{Since our initial depth is obtained by dilating sparse measurement points~\cite{ku2018defense}, surrounding points of most pixels are inaccurate.}
Therefore, at the beginning of the update process, we employ small scale cross-guided features to lead the TDU to predict a large scope, which obtains some long-range but accurate points as neighbors.
As the depth map becomes more refined, larger scale cross-guided features are sequentially used to pay more attention to shorter-range neighbors.
Due to the elegant recurrent strategy, our LRRU only requires four iterations to achieve satisfactory results.

Extensive experiments on both indoor and outdoor datasets verify the performance of our method.
Furthermore, we conduct comprehensive ablation studies to demonstrate the effectiveness of each component. 
Lastly, we extend our network framework to the depth-only case.

Our main contributions are summarized as: 
\vspace{-1.3mm}
\begin{itemize}
\setlength\itemsep{-1.3mm}
\item We propose a novel lightweight deep network architecture for depth completion, which pre-fills the sparse depth map and iteratively updates it by the proposed Target-Dependent Update (TDU) unit.
\item We propose a long-short range recurrent strategy, which dynamically adjusts kernel scopes during the update process to obtain long-to-short range dependencies.
\item As shown in \figref{fig:overview} and \tabref{tab:overview}, our four LRRU variants achieve state-of-the-art performance across different parameter regimes. Especially, the LRRU-Base model outperforms SOTA methods on NYUv2~\cite{Silberman12} and ranks 1st on the KITTI benchmark~\cite{chen2018estimating} at the time of submission.
\end{itemize}

\begin{table}[!t]
\centering
\footnotesize
\setlength{\tabcolsep}{0.8mm}
\caption{\textbf{The configurations of four LRRU variants}, which are obtained by adjusting the number of channels in different stages of the cross-guided feature extraction network.}
\label{tab:overview}
\begin{tabular}{@{}lcccccc@{}}
\hline
\multirow{2}{*}{Models} & \multicolumn{5}{c}{Number of channels}               & \multirow{2}{*}{~Params.} \\ \cline{2-6}
                        & stage1 & stage2 & stage3 & stage4 & stage5 &                             \\ 
\hline \hline
\textbf{LRRU-Mini~}  & 8   & 16   & 32  & 32  & 32  & 0.3 M      \\
\textbf{LRRU-Tiny~}  & 16  & 32   & 64  & 64  & 64  & 1.3 M      \\
\textbf{LRRU-Small~} & 32  & 64   & 128 & 128 & 128 & 5 M        \\
\textbf{LRRU-Base~}  & 64  & 128  & 256 & 256 & 256 & 21 M       \\ \hline
\end{tabular}
\vspace{-4mm}
\end{table}

\section{Related Work}\label{sec::relatedwork}

\noindent\textbf{Depth Completion.}
In the deep learning era, the straightforward depth completion methods~\cite{uhrig2017sparsity,zhao2021distance,wang2022ral} employ various network structures to directly predict dense depth maps.
Since the RGB images contain rich texture and semantic information, which are critical to recover the structure details of depth maps, many popular methods~\cite{ma2018sparse,MaCK19,qiu2019deeplidar,liu2021fcfr,HuWLNFG21} fuse the information of RGB images and sparse depth maps to boost the depth completion.
Ma~\etalcite{ma2018sparse} propose to concatenate the depth maps and the RGB images to form a 4D tensor, which is known as ``early-fusion''.
To reduce the gap between different modalities, some works\cite{MaCK19,qiu2019deeplidar,liu2021fcfr,HuWLNFG21} propose a ``later-fusion'' method, which extracts the feature of the RGB images and the sparse depth maps separately, and feeds the fusion features into the network. 
Further, Tang~\etalcite{tang2019learning} propose a feature fusion module based on the guided dynamic convolutional network~\cite{he2012guided,jia2016dynamic} to better utilize the guidance feature of RGB images.
In addition, some works~\cite{lee2021depth, wong2021learning, wong2021unsupervised} propose to first densify the sparse depth by classical approaches~\cite{ku2018defense,wong2020unsupervised}, and then learn a residual of the initial depth approximation.
However, existing methods usually employ massive parameters and computation to obtain good results.
A lightweight and efficient network architecture is lack.

\begin{figure*}[!t]
	\centering
	\includegraphics[width=0.87\textwidth]{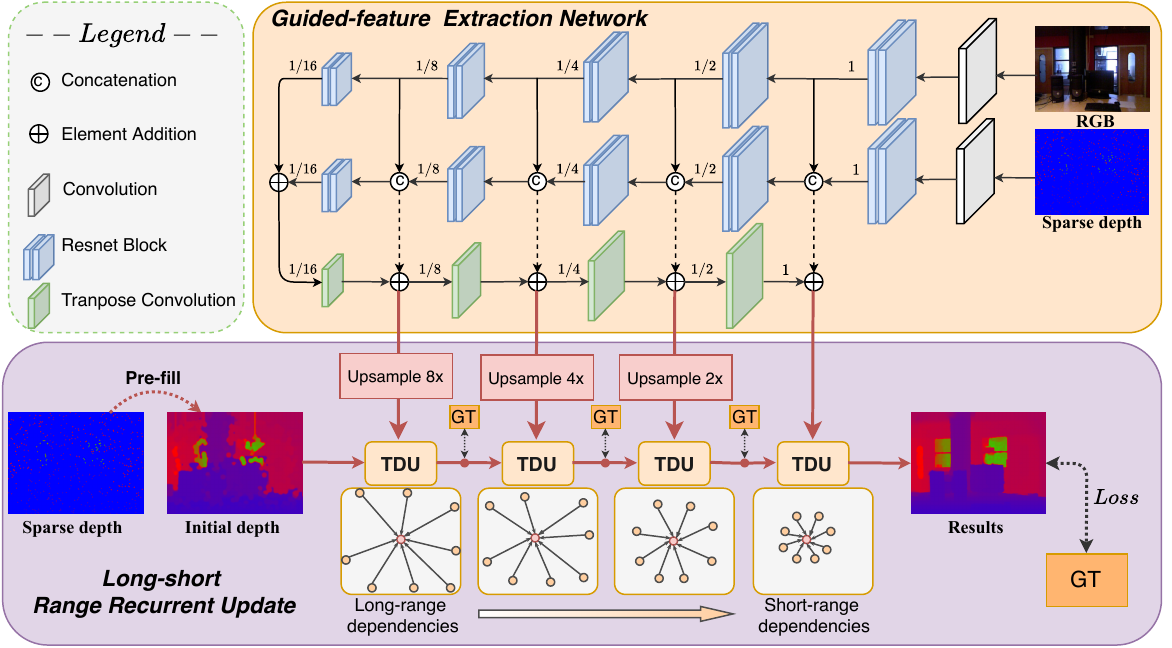}
	\caption{\textbf{The Long-short Range Recurrent Updating (LRRU) network}, which extracts the cross-guided features from RGB images and sparse depth map, and then iteratively updates the pre-filled depth map by \textbf{the proposed Target-Dependent Update (TDU) unit} according to \textbf{our long-short range recurrent strategy}.}
	\label{fig:method}
	\vspace{-4mm}
\end{figure*}

\noindent\textbf{Spatial Propagation Networks.}
The depth maps predicted by the direct-regression methods suffer from blur effect and distortion of object boundaries~\cite{cheng2019cspn}.
To address this issue, some works propose a series of spatial propagation networks (SPNs)~\cite{cheng2019cspn,cheng2020cspn++,xu2020deformable,park2020non,HuWLNFG21,lin2022dynamic,liu2022graphcspn}, which iteratively update the output of direct-regression methods by aggregating the reference and neighbor pixels.
The original SPN~\cite{liu2017learning} updates each pixel by three adjacent pixels from the previous row or column.
The serial update process is performed in four directions individually, and the results are combined by max-pooling.
To make the update process more efficient, Cheng~\etalcite{cheng2019cspn} propose the convolutional spatial propagation network (CSPN), which updates all pixels simultaneously within a fixed-local neighbors.
However, the fixed-local neighborhood configuration will introduce irrelevant points.
Furthermore, CSPN++~\cite{cheng2020cspn++} assembles the results obtained by using different kernel sizes.
DSPN~\cite{xu2020deformable} and NLSPN~\cite{park2020non} predict a non-local neighborhood by learning the offsets to the regular grid, while the difference between them is that DSPN obtains the kernel weights by calculating the similarity between features, while NLSPN learns it by the network.
Despite the improved SPN-based methods providing more flexibility in selecting neighbors, they employ fixed kernel weights during the update process, which limits the representation capability of SPN.
To alleviate the problem, DySPN~\cite{lin2022dynamic} gives variable weights to neighbors with different distances through learned attention maps.
GraphCSPN~\cite{liu2022graphcspn} leverages graph neural networks (GNN) to integrate 3D information into the update process.
However, existing SPNs still use fixed neighbors and are not able to dynamically adjust them during the update process.

\section{Method}\label{sec::method}
Given a sparse depth map, we first densify it by a simple non-learning method~\cite{ku2018defense}.
Then, according to the long-short range recurrent strategy (described in \secref{sec:iteration}), our method iteratively updates the initial depth map through the target-dependent update unit (described in \secref{sec:update}) to obtain accurate and dense depth map.
In \secref{sec:detail}, we provide implementation details of our method. 
For the convenience of description, we use \textbf{the target depth} (denoted as $\hat{D}^{\mathbf{t}}$ ) to refer to the depth map to be updated in the $t$-th update.
Thus, $\hat{D}^{\mathbf{1}}$ represents the initial map obtained by \cite{ku2018defense}.
Moreover, $\hat{D}^{\mathbf{t+1}}$ denotes the updated result.

\subsection{Target-dependent Update Unit}
\label{sec:update}

\note{The proposed Target-Dependent Update (TDU) unit updates the target depth map by learned spatially-variant kernels, which model neighbors and their affinities of each pixel.
To avoid irrelevant neighbors brought by the fixed-local neighborhood configuration, our TDU employs fully convolutional networks to predict the kernel weights and sampling position of neighbors as~\cite{park2020non}, where the sampling position is obtained by learning offsets to the regular grid.
However, direct supervisory information for the weights and offsets is typically not available, which often results in the training instability.
To overcome the instability, we use the features from RGB images and sparse depth maps to guide our TDU to obtain relevant neighbors, motivated by rich structure details in RGB images and the accurate scene depth information in sparse depth maps.
Since the dense RGB image and the sparse depth map belong to different modalities, we employ a dual-encoder network similar to~\cite{tang2019learning}, which uses two separate sub-networks to extract the features of RGB images and sparse depth maps respectively, and fuses them at multiple scales.
However, if only the features from RGB and sparse depth maps are used to guide the TDU, the update operation is independent of the content of the depth map to be updated, which may lead to sub-optimal solutions, especially when the initial depth map is not directly regressed from RGB images and sparse depth maps.
Therefore, in addition to RGB images and sparse depth maps, we propose to extract the features from the target depth map itself to guide our TDU.
We refer to the features form RGB images and sparse depth maps as the cross-guided features, and the features from the target depth map as the self-guided features.  
As shown in \equref{equ:feature}, the cross-guided features $F_{Cross-guided}$ are extracted from the input RGB image $I$ and sparse depth map $S$ by the feature extraction network $f_{\theta}$, and the self-guided features $F_{Self-guided}$ are obtained from the target depth map $\hat{D}^{\mathbf{t}}$ through a convolutional layer $f_{\psi}$.}
\begin{equation}
 F_{Cross-guided} = f_{\theta}(I, S), ~  F_{Self-guided} = f_{\psi}(\hat{D}^{\mathbf{t}}).
\label{equ:feature}
\end{equation}

\begin{figure}[!t]
	\centering
	\includegraphics[width=0.47\textwidth]{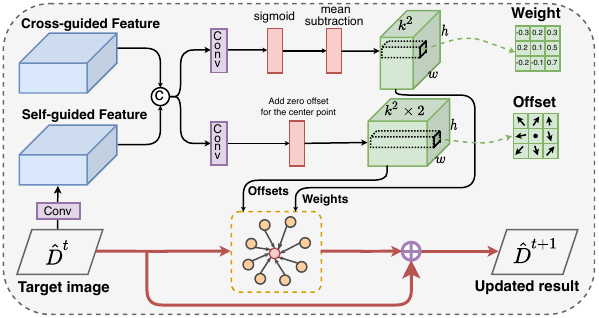}
	\caption{\note{\textbf{The Target-Dependent Update (TDU) unit} updates the target map by learning its residual map, where the cross-guided features from guidance images guide TDU to obtain relevant neighbors and the self-guided features from target maps ensure TDU adjust adaptively based on the content of the target map.}
	}
	\label{fig:update_block}
	\vspace{-4mm}
\end{figure}

\noindent\textbf{Weights and Offsets Regression.}
As shown in \figref{fig:update_block}, our TDU first concatenates the cross-guided features and the self-guided features, and then learns the weight and offset feature map by two individual $1\times 1$ convolutional layers.
To make the weights and offsets converge quickly, we add some restrictions to regulate their behaviors and guide the learning process.
Concretely, the weight feature map has $k^{2}$ channels, where $k$ is the kernel size and set to 3 in this paper.
We apply a sigmoid layer to make the weight larger than zero and smaller than one.
In addition, we subtract the mean value from the output of the sigmoid layer to make the sum of the weight to be zero, which performs like a a high-pass filter~\cite{kim2021def}.
The offset feature map has $2k^{2}$ channels, which represents the deviation of the sampling point from locations on a regular grid in the $x$ and $y$ directions.
However, to ensure that each reference pixel participates in its own update process, we first predict an offset feature map with $2(k^{2}-1)$ channels, and then insert zero into the center of the offset feature map channel~\cite{park2020non}.

\noindent\textbf{Residual Connection.}
We observe that the input and output map of the update unit are highly correlated, and share the low-frequency information.
Therefore, unlike existing SPNs~\cite{cheng2019cspn,cheng2020cspn++,xu2020deformable,park2020non,HuWLNFG21,lin2022dynamic,liu2022graphcspn} that directly predict the updated depth map, we propose to learn a residual image of the target depth map to enhance structural details and suppress noise.
Given the learned weights and sampling offsets, as shown in \equref{equ:residual}, the residual image $\Delta \hat{D}_{\mathbf{p}}^{\mathbf{t}}$ at position $\mathbf{p} = (x, y)$ is obtained by a weighted average.
\begin{equation}
\!\Delta \hat{D}_{\mathbf{p}}^{\mathbf{t}} =\!\!\!\sum_{\mathbf{q} \in \mathcal{N}(\mathbf{p})}\!\!\!\mathbf{W}_{\mathbf{pq}}(F_{Cross-guided}, F_{Self-guided}) \hat{D}_{\mathbf{q}}^{\mathbf{t}}.
\label{equ:residual}
\end{equation}

In \equref{equ:residual}, $\mathcal{N}(\mathbf{p})$ denotes the set of neighbors of the position $\mathbf{p}$. 
Since the offsets are normally fractional, we use the bilinear interpolation to sample local four points as \cite{dai2017deformable} and \cite{zhu2019deformable}.
The filter weights $\mathbf{W}$ are predicted from the cross-guided and self-guided features.
We aggregate depth values from the sparsely chosen locations with the learned weights.
Then, we add the residual image to the target depth map as \equref{equ:add} to obtain the updated depth map $\hat{D}^{\mathbf{t+1}}$.
\begin{equation}
\hat{D}^{\mathbf{t+1}} = \hat{D}^{\mathbf{t}} + \Delta \hat{D}^{\mathbf{t}}.
\label{equ:add}
\end{equation}

\begin{figure}[!t]
	\centering
	\includegraphics[width=0.47\textwidth]{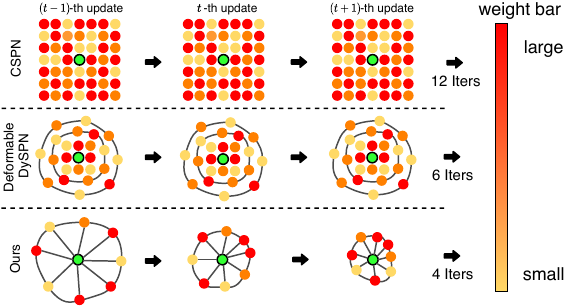}
	\caption{\textbf{The Long-short Range Recurrent Strategy}, which dynamically adjust the kernel scope from large to small during the update process, while existing SPNs keep it unchanged.}
		\vspace{-3mm}
	\label{fig:recurrent_scheme}
\end{figure}

\subsection{Long-short Range Recurrent Update Strategy}
\label{sec:iteration}

During the update process, we need an effective recurrent update strategy to enable the TDUs to obtain appropriate kernel weights and scopes for their respective target.
Specifically, for the initial depth map obtained by a non-learning method~\cite{ku2018defense}, only a few available sparse measurement points and their surrounding points have high accuracy, the surrounding points of most pixels are inaccurate~\cite{wang2022ral}.
Therefore, at the beginning of the update process, we should employ a large kernel scope to obtain some long-range but accurate points as neighbors.
As the depth map becomes more refined, the kernel scope should be gradually narrowed to pay more attention to short-range points, as they are generally more relevant to the reference point.
However, existing recurrent update strategies used by SPNs are not flexible enough to meet the above needs.
For example, CSPN~\cite{cheng2019cspn} and NLSPN~\cite{park2020non} use fixed kernel weights and scopes during the update process, which not only limits the representation capability of SPNs, but also requires massive iterations to obtain long-range dependencies.
Although CSPN++~\cite{cheng2020cspn++} and DySPN~\cite{lin2022dynamic} alleviate this issue by the model ensemble and attention mechanism, their kernel scopes remain unchanged during the update process, even in the deformable DySPN.

\begin{figure}[!t]
\centering
\begin{minipage}[b]{.47\textwidth}
\centerline{\includegraphics[width=\linewidth]{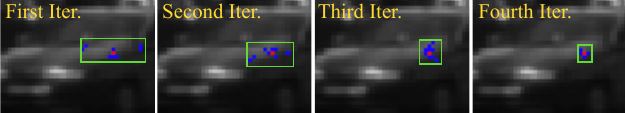}}
\caption{Typical examples to illustrate that large-to-small kernel scopes are dynamically adjusted to capture long-to-short range dependencies during the iterative update process (the red points denote the reference pixel and blue points denote the neighbors).}\label{offset-a}
\vspace{2mm}
\end{minipage}\qquad
\begin{minipage}[b]{.47\textwidth}
\centerline{\includegraphics[width=\linewidth]{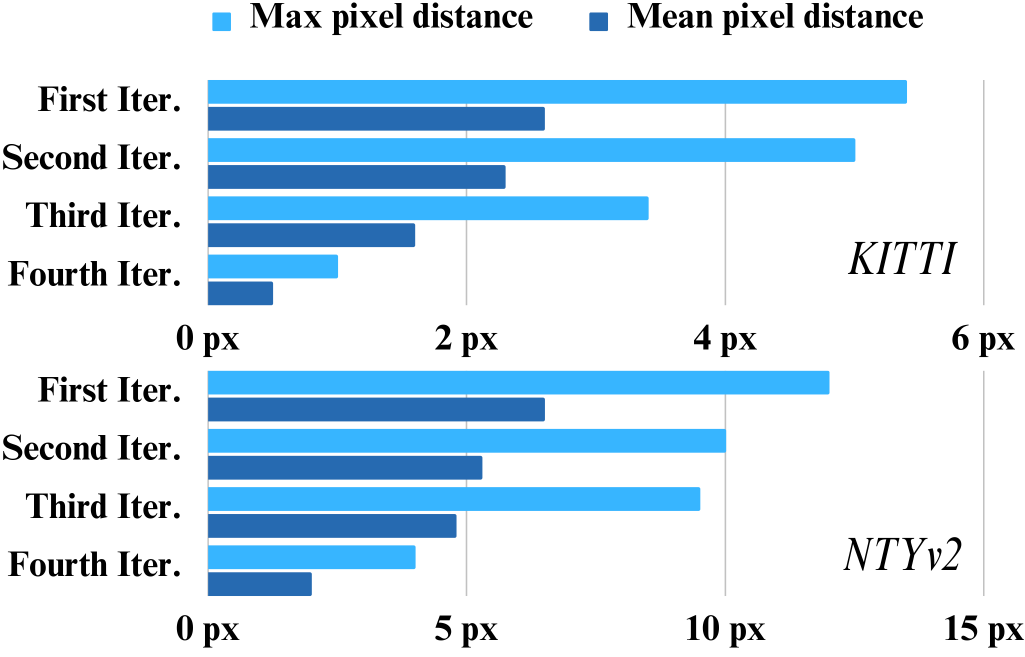}}
\caption{Analysis of the max and mean pixel distances from neighbors to the reference pixel on  KITTI and NYUv2 dataset.}\label{offset-b}
\vspace{-4mm}
\end{minipage}
\end{figure}

In this paper, we propose a long-short range recurrent update strategy that is shown in \figref{fig:recurrent_scheme}.
The parameters of each TDU, including the kernel weights and sampling positions of neighbors, are learned by considering the cross-guided and self-guided features.
We observe that when TDUs of the iterative update process are guided by the cross-guided features of different scales respectively, the TDU guided by smaller scale cross-guided features will adaptively learn to obtain the neighbors with a relatively large scope, and vice versa.
We consider this is due to the different receptive fields of the cross-guided features at different scales.
Building upon the above observations, we employ the $1/8$-scale cross-guided feature map to guide the TDU of the first iteration to obtain the neighbors with a large scope.
In subsequent iterations, the TDU gradually uses the larger scale cross-guided features, such as ${1}/{4}$-scale, ${1}/{2}$-scale, and full-scale, to obtain the neighbors with smaller scopes.
\figref{offset-a} and \figref{offset-b} demonstrate the kernel scope changes from large to small during the iterative update process on the KITTI and NYUv2 dataset.
Since the proposed recurrent update strategy is highly flexible, we can achieve satisfactory results by using fewer iterations and neighbors.

\subsection{Implementation Details}
\label{sec:detail}

\noindent\textbf{Network Architecture.}
Our network architecture, shown in \figref{fig:method} consists of the cross-guided feature extraction network and the long-short range recurrent update module.
The cross-guided feature extraction network employs two sub-networks, the depth encoder and the RGB encoder, to extract features from the sparse depth map and corresponding RGB image respectively. 
The extracted multi-scale RGB features are injected into the depth encoder to fully integrate the information from 
different modalities.
Then, a decoder network is used to learn the residual of the fused multi-scale features.
The cross-guided features are first upsampled to the same resolution as the initial depth map, and utilized in the TDUs of the iterative update process.

\noindent\textbf{Loss Function.}
We supervise our network through $\mathcal{L}_1$ and $\mathcal{L}_2$ distance between the output and the ground truth $D_{gt}$ over results of each iteration, $\{\hat{D}^{\mathbf{2}}, \cdots, \hat{D}^{\mathbf{N}}\}$, with exponentially increasing weights. The loss is defined in Eq.~\eqref{equ:l2}.
\begin{equation}
	\mathcal{L}=\sum_{\sigma=1}^{2} \sum_{i=2}^{N} \gamma^{N-i}\| (\hat{D}^{i} - D_{gt}) \odot \mathbf{1}_{\{D_{gt} > 0\}}  \|^{\sigma}, \label{equ:l2}
\end{equation}
where $\mathbf{1}$ indicates whether there is a value in the ground truth, and $\odot$ denotes the element-wise multiplication.
We set $\gamma = 0.8$ in our experiments.

\noindent\textbf{Training Details.} 
We employ PyTorch~\cite{paszke_pytorch_NIPS_2019} to implement our model, which is trained and tested with GeForce RTX 3090 GPUs.
All models are initialized from scratch with random weights.
During training, we employ the Adam optimizer with a batch size of 8.
We set $\beta_{1} = 0.9$, $\beta_{2} = 0.999$, weight decay is $10^{-6}$, and the total number of epochs is 45.
The initial learning rate is $10^{-3}$, and the learning rate remains unchanged for the first 15 epochs and decreases by $50\%$ every 5 epochs.

\section{Experiments}\label{sec::experiments}

\begin{figure*}[!t]
	\centering
	\includegraphics[width=0.87\textwidth]{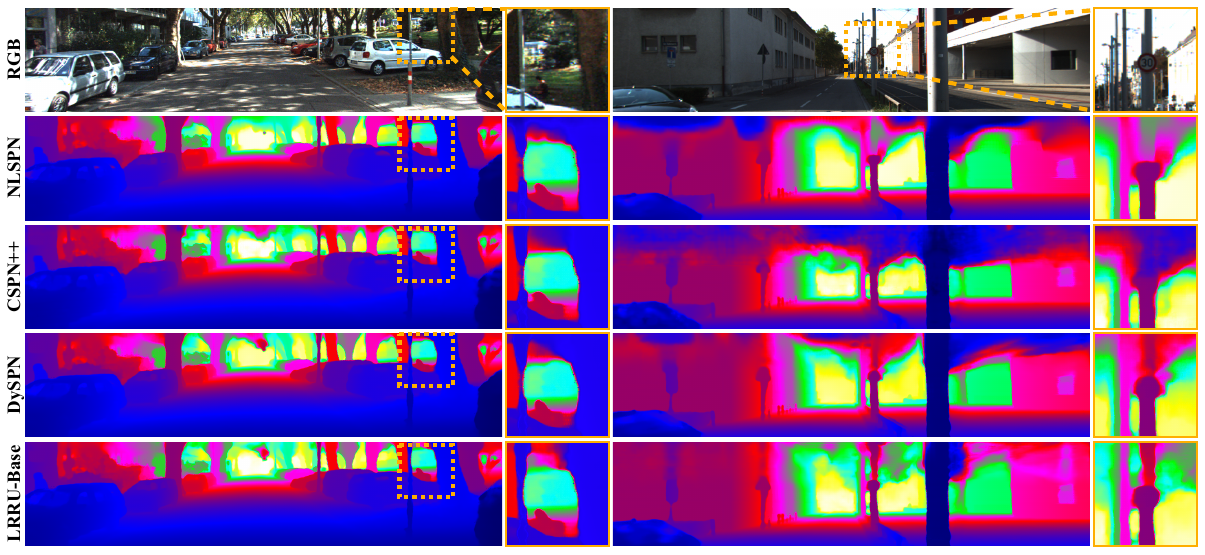}
	\vspace{-2mm}
	\caption{Qualitative comparison with SOTA methods on KITTI test dataset. From top to bottom are RGB images, dense depth maps predicted by NLSPN~\cite{park2020non}, CSPN++~\cite{cheng2020cspn++}, DySPN~\cite{lin2022dynamic} and Ours, respectively. We zoom in some representative areas for detailed comparison.}
	\label{fig:qualitative}
	\vspace{-4mm}
\end{figure*}

\subsection{Datasets and Metrics}

\noindent\textbf{KITTI Dataset~\cite{Geiger2012CVPR}.}
The KITTI dataset, a popular real-world autonomous driving dataset, consists of sparse depth maps projected from raw LiDAR scans and corresponding RGB images.
It contains 86k frames for training, 1k selected frames for validation, and 1k frames without ground truth that need to be tested on the KITTI online benchmark.

\noindent\textbf{NYUv2 Dataset~\cite{Silberman12}.}
The NYUv2 dataset consists of RGB and depth images acquired from 464 different indoor scenes. 
Following the standard setting~\cite{MaCK19}, we train the model with 50K images sampled from the training set, and test it on 654 official labeled images.
For both train and test datasets, the original images of size $640\times 480$ are downsampled to half and then center-cropped to $304\times 228$. 

\noindent\textbf{Evaluation Metrics.}
Following exiting depth completion methods~\cite{MaCK19,park2020non,tang2019learning}, we employ Root Mean Squared Error (RMSE),  Mean Absolute Error (MAE), inverse RMSE (iRMSE), inverse MAE (iMAE), mean absolute relative error (REL), and percentage of pixels satisfying $\delta_{\tau}$ for quantitative evaluation.

\subsection{Comparison with SOTA Methods on KITTI}

\noindent\textbf{Quantitative Comparison.}
\tabref{tab:sota} reports the quantitative evaluation of our methods and state-of-the-art (SOTA) depth completion methods on the KITTI benchmark, which ranks all methods according to the RMSE value.
Since directly predicting dense depth maps from sparse depth maps and RGB images is a difficult ``ill-posed'' task, most direct-regression methods, such as S2D~\cite{MaCK19}, DeepLiDAR~\cite{qiu2019deeplidar}, GuideNet~\cite{tang2019learning},  ACMNet\cite{zhao2021adaptive}, PENet~\cite{HuWLNFG21}, and RigNet\cite{yan2021rignet}, generally require massive stacked filters and layers to learn robust features.
They usually have a large number of model parameters and a relatively long inference time, while the performances of the method are difficult to achieve the best of the time~\cite{Geiger2012CVPR}.
Even the efficiency-oriented depth completion methods, such as excellent GAENet~\cite{chen2022depth} and FuseNet\cite{qiu2019deeplidar}, are not able to achieve a good balance between model parameters, inference time and performance.
To improve the performance of direct-regression methods, a series of SPNs are proposed to refine the output of these methods, such as CSPN~\cite{cheng2019cspn}, NLSPN~\cite{park2020non}, CSPN++~\cite{cheng2020cspn++}, and DySPN\cite{lin2022dynamic}.
The results show that the supplemental refinement modules improve the performance of the method, but also requires longer inference time.

\begin{table}[!t]
    \small
    \setlength{\tabcolsep}{0.8mm}
        \centering
        \caption{Quantitative comparison with state-of-the-art (SOTA) methods on the KTTTI benchmark. 
        The best and second-best results are highlighted in \redfont{red} and \bluefont{blue} colors
        , respectively.}
        \label{tab:sota}
        \resizebox{0.5\textwidth}{!}{%
        \begin{tabular}{lrccccc}
        \hline
        {Methods} &
          {RMSE}  {[}mm{]} &
          {MAE}  {[}mm{]} &
          {iRMSE}  {[}1/km{]} &
          {iMAE}  {[}1/km{]} \\ \hline \hline
        CSPN\cite{cheng2019cspn}                    & 1019.64               & 279.46              & 2.93                 & 1.15              \\
        S2D\cite{MaCK19}                         & 814.73               & 249.95               & 2.80                 & 1.21              \\
        GAENet\cite{chen2022depth}                   & 773.90               & 231.29               & 2.29                 & 1.08                 \\
        DeepLiDAR\cite{qiu2019deeplidar}                     & 758.38               & 226.50               & 2.56                 & 1.15                 \\
        FuseNet\cite{qiu2019deeplidar}                    & 752.88               & 221.19               & 2.34                 & 1.14                 \\
        MDANet\cite{MDANet}                        & 738.23               & 214.99               & \textcolor{black}{{2.12}}                & 0.99                 \\
        GuideNet\cite{tang2019learning}                       & 736.24               & 218.83               & 2.25                 & 0.99                 \\
        CSPN++\cite{cheng2020cspn++}               & 743.69        & 209.28    & 2.07 &           0.90     \\
          NLSPN\cite{park2020non}                          & 741.68               & 199.59              & 1.99                & 0.84       \\
        ACMNet\cite{zhao2021adaptive}                        & 744.91               & 206.09               & \textcolor{black}{2.08}                 & 0.90                 \\
        PENet\cite{HuWLNFG21}                       & 730.08               & 210.55               & 2.17                 & 0.94                 \\
        RigNet\cite{yan2021rignet}                  & \textcolor{black}{{712.66}}              & \textcolor{black}{{203.25}}              & \textcolor{black}{2.08}                & 0.90     \\ 
        DySPN\cite{lin2022dynamic}                      & \bluefont{709.12}        & \bluefont{192.71}   & \bluefont{1.88}     & \bluefont{0.82}    \\ \hline
        LRRU-Mini      & {774.43} & \multicolumn{1}{c}{210.87} & \multicolumn{1}{c}{2.21} & \multicolumn{1}{c}{0.90} \\
        LRRU-Tiny      & {738.86} & \multicolumn{1}{c}{200.28} & \multicolumn{1}{c}{2.04} & \multicolumn{1}{c}{0.85} \\
        LRRU-Small        & {717.50} & \multicolumn{1}{c}{197.72} & \multicolumn{1}{c}{1.96} & \multicolumn{1}{c}{\textcolor{black}{{0.85}}} \\
        LRRU-Base        & {\textcolor{red}{\textbf{696.51}}} & \multicolumn{1}{c}{\textcolor{red}{\textbf{189.96}}} & \multicolumn{1}{c}{\redfont{1.87}} & \multicolumn{1}{c}{\textcolor{red}{\textbf{0.81}}} \\ \hline
        \end{tabular}
        }
        
    \vspace{-4mm}
\end{table}

\begin{table*}[!t]
\centering
\footnotesize
\setlength{\tabcolsep}{0.8mm}
\caption{Ablation studies of the proposed LRRU on the KITTI validation dataset. ``x-scale'' denotes the scale of the cross-guided feature, and ``$\leftarrow$'' indicates that LRRU uses 1/8-scale, 1/4-scale, 1/2-scale, and full-scale cross-guided features sequentially during the update process. The results of our four LRRU variants are highlighted with a gray background.
The lower result values represent better performance.
}
\vspace{1mm}
\label{tab:ablation}
\begin{tabular}{@{}cccccccccccccc@{}}
\hline
\multirow{2}{*}{} &
  \multirow{2}{*}{Models} &
  \multicolumn{5}{c}{Long-short Range Recurrent Update Strategy} &
  \multicolumn{5}{c}{Target-Dependent Update Unit} &
  \multicolumn{2}{c}{Results} \\ \cmidrule(l){3-7} \cmidrule(l){8-12} \cmidrule(l){13-14}
 &
   &
  Full-scale &
  1/2-scale &
  1/4-scale &
  1/8-scale &
  iterative &
  \begin{tabular}[c]{@{}c@{}}w/o kernel\\ prediction \end{tabular} &
  \begin{tabular}[c]{@{}c@{}}w/o self-guided\\  feature \end{tabular} &
  \begin{tabular}[c]{@{}c@{}}w/o \\ Res.\end{tabular} &
  \begin{tabular}[c]{@{}c@{}}w/o mean \\ subtraction\end{tabular} &
  \begin{tabular}[c]{@{}c@{}}w/o fix \\ reference\end{tabular} &
 RMSE[mm]&
  MAE[mm] \\ \hline \hline
\multicolumn{1}{c|}{(a)} &
  \multicolumn{1}{c|}{Baseline} &
   &
   &
   &
   &
   &
   &
   &
   &
   &
  \multicolumn{1}{c|}{} &
   1027.1&302.1
   \\ \hline
\multicolumn{1}{c|}{(b)} &
  \multicolumn{1}{c|}{\multirow{6}{*}{Mini}} &
  $\checkmark$ &
   &
   &
   &
   &
   &
   &
   &
   &
  \multicolumn{1}{c|}{} &
   819.4&228.9
   \\
\multicolumn{1}{c|}{(c)} &
  \multicolumn{1}{c|}{} &
   &
  $\checkmark$ &
   &
   &
   &
   &
   &
   &
   &
  \multicolumn{1}{c|}{} &
   823.6&228.4
   \\
\multicolumn{1}{c|}{(d)} &
  \multicolumn{1}{c|}{} &
   &
   &
  $\checkmark$ &
   &
   &
   &
   &
   &
   &
  \multicolumn{1}{c|}{} &
   825.2&229.4
   \\
\multicolumn{1}{c|}{(e)} &
  \multicolumn{1}{c|}{} &
   &
   &
   &
  $\checkmark$ &
   &
   &
   &
   &
   &
  \multicolumn{1}{c|}{} &
   838.3&237.3
   \\
\multicolumn{1}{c|}{(f)} &
  \multicolumn{1}{c|}{} &
  $\checkmark$ &
  $\checkmark$ &
  $\checkmark$ &
  $\checkmark$ &
   &
   &
   &
   &
   &
  \multicolumn{1}{c|}{} &
   812.4&226.3
    \\
   \multicolumn{1}{c|}{(g)} &
  \multicolumn{1}{c|}{} &
  $\checkmark$ &
  $\checkmark$ &
  $\checkmark$ &
  $\checkmark$ &
  $\rightarrow$ &
   &
   &
   &
   &
  \multicolumn{1}{c|}{} &
   844.5&236.4
   \\
\myrowcolour \multicolumn{1}{c|}{(h)} &
  \multicolumn{1}{c|}{} &
  $\checkmark$ &
  $\checkmark$ &
  $\checkmark$ &
  $\checkmark$ &
  $\leftarrow$ &
   &
   &
   &
   &
  \multicolumn{1}{c|}{} &
   800.9&218.9
   \\ \hline
\multicolumn{1}{c|}{(i)} &
  \multicolumn{1}{c|}{\multirow{5}{*}{Mini}} &
  $\checkmark$ &
  $\checkmark$ &
  $\checkmark$ &
  $\checkmark$ &
  $\leftarrow$ &
  $\checkmark$ &
   &
   &
   &
  \multicolumn{1}{c|}{} &
   812.2&219.7
   \\
\multicolumn{1}{c|}{(j)} &
  \multicolumn{1}{c|}{} &
  $\checkmark$ &
  $\checkmark$ &
  $\checkmark$ &
  $\checkmark$ &
  $\leftarrow$ &
   &
  $\checkmark$ &
   &
   &
  \multicolumn{1}{c|}{} &
   817.5&223.1
   \\
\multicolumn{1}{c|}{(k)} &
  \multicolumn{1}{c|}{} &
  $\checkmark$ &
  $\checkmark$ &
  $\checkmark$ &
  $\checkmark$ &
  $\leftarrow$ &
   &
   &
  $\checkmark$ &
   &
  \multicolumn{1}{c|}{} &
   803.4&221.5
   \\
\multicolumn{1}{c|}{(l)} &
  \multicolumn{1}{c|}{} &
  $\checkmark$ &
  $\checkmark$ &
  $\checkmark$ &
  $\checkmark$ &
  $\leftarrow$ &
   &
   &
   &
  $\checkmark$ &
  \multicolumn{1}{c|}{} &
   804.0&224.8
   \\
\multicolumn{1}{c|}{(m)} &
  \multicolumn{1}{c|}{} &
  $\checkmark$ &
  $\checkmark$ &
  $\checkmark$ &
  $\checkmark$ &
  $\leftarrow$ &
   &
   &
   &
   &
  \multicolumn{1}{c|}{$\checkmark$} &
   799.4&222.2
   \\ \hline
\myrowcolour \multicolumn{1}{c|}{(n)} &
  \multicolumn{1}{c|}{Tiny} &
  $\checkmark$ &
  $\checkmark$ &
  $\checkmark$ &
  $\checkmark$ &
  $\leftarrow$ &
   &
   &
   &
   &
  \multicolumn{1}{c|}{} &
   761.5&207.6
   \\
\myrowcolour \multicolumn{1}{c|}{(o)} &
  \multicolumn{1}{c|}{Small} &
  $\checkmark$ &
  $\checkmark$ &
  $\checkmark$ &
  $\checkmark$ &
  $\leftarrow$ &
   &
   &
   &
   &
  \multicolumn{1}{c|}{} &
   741.3&201.8
   \\
\myrowcolour \multicolumn{1}{c|}{(p)} &
  \multicolumn{1}{c|}{Base} &
  $\checkmark$ &
  $\checkmark$ &
  $\checkmark$ &
  $\checkmark$ &
  $\leftarrow$ &
   &
   &
   &
   &
  \multicolumn{1}{c|}{} &
   728.3&197.9
   \\ \hline
\end{tabular}
\vspace{-3mm}
\end{table*}

By contrast, our proposed method (LRRUs) is more effective.
Specifically, our smallest LRRU-Mini model achieves better results than some large-scale methods, such as S2D~\cite{MaCK19} and CSPN~\cite{cheng2019cspn}.
In addition, LRRU-Tiny model significantly improves the performance over LRRU-Mini by increasing only 1M parameters, which outperforms many excellent methods, such as DeepLiDAR~\cite{qiu2019deeplidar}.
The experimental results strongly demonstrate the effectiveness of our proposed method.
It is worth noting that our largest LRRU-Base model achieves the best results of the KITTI benchmark in all evaluation metrics, including RMSE, MAE, iRMSE and iMAE, which \textbf{ranks 1st} at the time of submission.
To the best of our knowledge, it is the first depth completion method that obtains an RMSE value below 700 mm.

\noindent\textbf{Qualitative Comparison.}
\figref{fig:qualitative} compares the dense depth maps predicted by our LRRU-Base and several popular SPNs-based methods, such as NLSPN~\cite{park2020non}, CSPN++~\cite{cheng2020cspn++}, and DySPN~\cite{lin2022dynamic}.
These SPNs-based methods use additional refinement modules, which are generally considered to have good visualization results.
However, as shown in the enlarged results, the dense depth maps predicted by these SPNs-based methods have depth discontinuity and blur effect in some areas, especially in thin structures.  
By contrast, our results have better structure details and are more accurate at the object boundaries.

\subsection{Ablation Studies}
\label{sec:ablation}

To reduce the training time, the ablation experiments are conducted on our smallest \textbf{LRRU-Mini model}.

\noindent\textbf{Network Architectures.}
Existing deep-learning methods based on direct regression rely heavily on the representation capability of the network.
When the network size decreases, the method performance drops significantly.
We select the method that employs the cross-guided feature extraction network to directly predict dense depth maps as the baseline.
\tabref{tab:ablation} (a) shows that the results of the baseline method has a large error, the RMSE is 1027.1 mm, and the MAE is 302.1 mm.
Unlike direct regression, we iteratively update the initial coarse depth map by the proposed TDU to obtain an accurate depth map.
As shown in \tabref{tab:ablation} (b)-(e), even if we only employ a TDU guided by the cross-guided feature of arbitrary scales to update the initial depth map once, the proposed method achieves better performance.
Specifically, compared to the baseline, the results obtained by using the full-scale cross-guided feature reduce RMSE and MAE by 207.7 mm and 73.2 mm, respectively.
It illustrates that our proposed network framework is more effective than direct regression when the network size is limited.
In addition, As shown in \tabref{tab:ablation} (f), when we use a TDU guided by the fusion cross-guided feature of various scales, the method performance is further improved.

\noindent\textbf{Long-short Range Recurrent Update Strategy.}
In this paper, we propose a long-short range recurrent update strategy to manipulate the update process.
By sequentially using small-to-large scale cross-guided feature to guide the TDU, our method dynamically obtains long-to-short range dependencies.
The results in \tabref{tab:ablation}~(h) show that the method using our proposed strategy achieves substantially improved performance.
Meanwhile, we report the results obtained by using the cross-guided features in the reverse order of ours, namely a short-to-long range recurrent update strategy.
The results in \tabref{tab:ablation} (g) show that the method using this strategy achieves poor results, which demonstrates the advantage of the proposed recurrent strategy.

\noindent\textbf{Target-Dependent Update Unit.}
The TDU explicitly builds the cross-guided feature and self-guided feature into update unit, and updates the depth map by learned spatially-variant kernels.
If the TDU predicts the residual map by direct regression rather than learning spatially-variant kernels, the results in \tabref{tab:ablation} (i) show that the method performance will decrease significantly.
Meanwhile, as shown in \tabref{tab:ablation} (j), TDU without using the self-guided feature also heavily harms the method results.
In addition, if the residual learning is not adopted, the performance shown in \tabref{tab:ablation} (k) will decrease.
As described in \secref{sec:update}, we perform many operations, including subtracting the mean value from the output of the sigmoid layer and fixing the reference pixel of the update unit, to regulate the behaviors of the weights and offsets and guide their learning process.
The results in \tabref{tab:ablation} (l) and (m) show that these restrictions are beneficial to improve the performance of the method.
It is worth noting that although fixing the reference pixel is not able to boost the quantitative results greatly, we empirically observe that it effectively avoids unreliable neighboring pixels.

\noindent\textbf{LRRU vs existing SPNs.}
To better show the advantages of our method, we employ the refinement modules of existing SPNs directly in our proposed framework.
We select the refinement module of NLSPN~\cite{park2020non} under its original setting for the comparison, which performs best among all open-source SPN-based methods.
The results in \tabref{tab:nlspn} show that our framework using existing refinement module directly performs worse than our method, which illustrate the effectiveness of our proposed method.

\begin{table}[!h]
\centering
\setlength{\tabcolsep}{1mm}
\vspace{-2mm}
\caption{The performance comparison of our proposed framework using the refinement module of this paper and existing SPNs.}
\label{tab:nlspn}
\resizebox{.47\textwidth}{!}{%
\begin{tabular}{lcccc}
\hline
Methods & RMSE {[}mm{]} & MAE {[}mm{]} & iRMSE {[}1/km{]} & iMAE {[}1/km{]} \\ \hline \hline
        LRRU-Mini (Ours)                 & 800.9              & 218.9              & 2.4                 & 0.9              \\
        LRRU-Mini(NLSPN~\cite{park2020non})    & 891.5               & 240.2             & 5.0                 & 1.1              \\ \hline
\end{tabular}%
}
\end{table}

\noindent\textbf{Loss Functions.}
The above ablation experiments are trained with $\mathcal{L}_2$ loss.
As shown in \tabref{tab:loss}, we test the combination of $\mathcal{L}_1$ and $\mathcal{L}_2$ losses that improves MAE performance while decreases RMSE performance slightly.
For a balance between the RMSE and MAE, we adopt the $\mathcal{L}_1 + \mathcal{L}_2$ loss to train the model on the KITTI test dataset.

\begin{table}[!h]
\centering
\setlength{\tabcolsep}{1mm}
\vspace{-2mm}
\caption{The performance comparison of using different loss.}
\label{tab:loss}
\resizebox{.47\textwidth}{!}{%
\begin{tabular}{lccccc}
\hline
Methods & Loss & RMSE {[}mm{]} & MAE {[}mm{]} & iRMSE {[}1/km{]} & iMAE {[}1/km{]} \\ \hline \hline
LRRU-Mini     & $\mathcal{L}_2+\mathcal{L}_1$           & 806.3              & 210.2              & 2.3                 & 0.9              \\
LRRU-Tiny      & $\mathcal{L}_2+\mathcal{L}_1$           & 763.8               & 198.9             & 2.1                 & 0.8              \\ 
LRRU-Small      & $\mathcal{L}_2+\mathcal{L}_1$           & 745.3               & 195.7             & 2.0                 & 0.8               \\ 
LRRU-Base       & $\mathcal{L}_2+\mathcal{L}_1$          & 729.5               & 188.8             & 1.9                 & 0.8               \\ 
\hline
\end{tabular}%
}
\end{table}

\noindent\textbf{Hardware cost comparison.}
\tabref{tab:runtime} compares the hardware cost of LRRUs and most open-source methods on the KITTI validate dataset (the resolution of the input image is $356*1216$), including the model parameters, GPU memory, and runtime.
The experiment is conducted on the same hardware (a GeForce RTX 3090 GPU).
Although our pre-filled method (IP-BASIC~\cite{ku2018defense}) needs extra 8ms CPU-time, the proposed LRRUs are more efficient.
Especially, our smallest LRRU-Mini model contains only 0.3M parameters and has a fast inference speed (38.3ms).

\begin{table}[!h]
\centering
\caption{ {Hareware cost comparison of KITTI validate dataset (the resolution of input images $356*1216$) on a 3090 GPU.} 
Our pre-filled method (IP-BASIC~\cite{ku2018defense}) needs extra 8ms CPU-time.}
\setlength{\tabcolsep}{1mm}
\label{tab:runtime}
\resizebox{0.45\textwidth}{!}{%
\begin{tabular}{@{}cccccccc@{}}
\hline
Methods             & Parameters & \multicolumn{1}{c}{GPU Memory} & GPU Time & CPU Time & Total Time \\ \hline \hline
MDANet\cite{MDANet}   & 3M     & \multicolumn{1}{c}{2092M}  & 89.4ms   & -        & 89.4ms  \\
CSPN\cite{cheng2019cspn}      & 26M    & \multicolumn{1}{c}{2350M}  & 152.7ms  & -        & 152.7ms \\
S2D\cite{MaCK19}       & 26M    & \multicolumn{1}{c}{2706M}  & 65.5ms   & -        & 65.5ms  \\
DeepLiDAR\cite{qiu2019deeplidar}  & 144M   & \multicolumn{1}{c}{3496M}  & 323.6ms  & -        & 323.6ms \\
NLSPN\cite{park2020non}     & 26M    & \multicolumn{1}{c}{2628M}  & 76.0ms   & -        & 76.0ms  \\
GuideNet\cite{tang2019learning}   & 74M    & \multicolumn{1}{c}{2074M}  & 51.3ms   & -        & 51.3ms  \\
PENet\cite{HuWLNFG21}     & 132M   & \multicolumn{1}{c}{2822M}  & 129.6ms  & -        & 129.6ms \\ \hline
LRRU-Mini          & \textbf{0.3M} & \multicolumn{1}{c}{\textbf{1448M}} & \textbf{30.3ms} & 8ms & \textbf{38.3ms} \\
LRRU-Tiny         & 1.3M   & \multicolumn{1}{c}{1672M}  & 41.4ms   & 8ms      & 49.4ms  \\
LRRU-Small        & 5M     & \multicolumn{1}{c}{2152M}  & 59.3ms   & 8ms      & 67.3ms  \\
LRRU-Base         & 21M           & \multicolumn{1}{c}{3170M}          & 117.4ms         & 8ms & 125.4ms         \\ \hline
\end{tabular}%
}
\end{table}

\noindent\textbf{Test-time augmentation (TTA).} The TTA is widely used in depth completion methods (such as GuideNet~\cite{tang2019learning}, ACMNet~\cite{zhao2021adaptive}, \textbf{\etal}) as a performance-enhancing strategy, but is rarely described.
In the inference stage, the model with TTA predicts the results of the original sample and the horizontally reversed sample at the same time, and averages their results as the final result.
\tabref{tab:depth-only} shows that the TTA strategy makes the model obtain lower RMSE, especially in the smallest LRRU-Mini model.

\begin{table}[!t]
\centering
\setlength{\tabcolsep}{1mm}
\vspace{-2mm}
\caption{Results of LRRU models on the KITTI validate dataset with test-time augmentation (TTA).}
\label{tab:tta}
\resizebox{.47\textwidth}{!}{%
\begin{tabular}{lccccc}
\hline
Methods & TTA & RMSE {[}mm{]} & MAE {[}mm{]} & iRMSE {[}1/km{]} & iMAE {[}1/km{]} \\ \hline \hline
LRRU-Mini     & False        & 806.3              & 210.2              & 2.3                 & 0.9              \\
LRRU-Tiny      & False          & 763.8               & 198.9             & 2.1                 & 0.8              \\ 
LRRU-Small      & False           & 745.3               & 195.7             & 2.0                 & 0.8               \\ 
LRRU-Base       & False          & 729.5               & 188.8             & 1.9                 & 0.8               \\ 
\hline
LRRU-Mini     & True        &     0.7943          & 208.5              & 2.3                 & 0.9              \\
LRRU-Tiny      & True          & 757.8               & 198.0             & 2.1                 & 0.8              \\ 
LRRU-Small      & True           & 739.2              & 195.0             & 2.0                 & 0.8               \\ 
LRRU-Base       & True          & 723.4               & 188.1             & 1.9                 & 0.8               \\ 
\hline
\end{tabular}%
}
\end{table}

\subsection{Experiments on fewer points}

\begin{figure}[!t]
	\centering
	\includegraphics[width=0.45\textwidth]{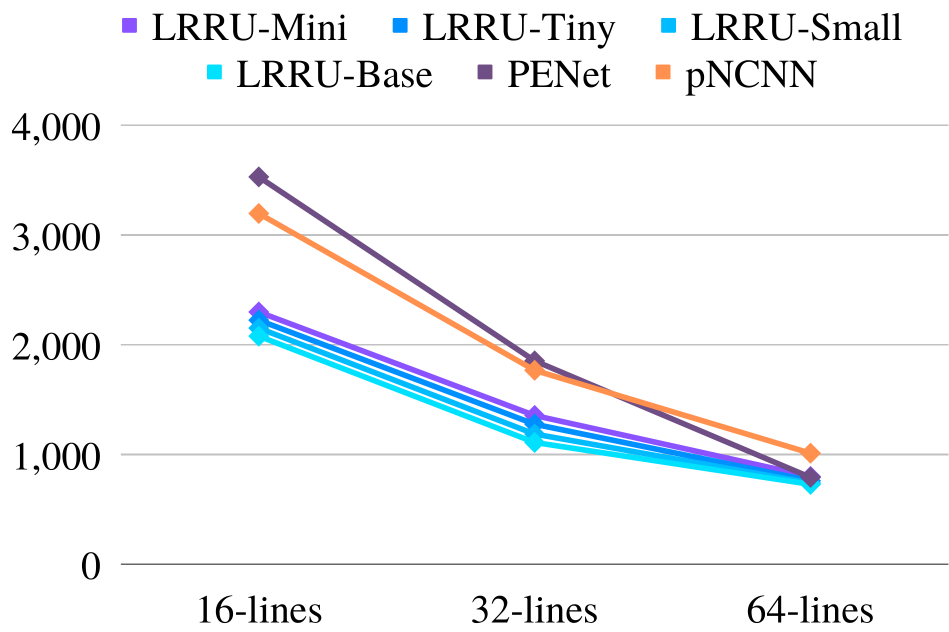}
	\caption{The performance (RMSE [mm]) of our methods, PENet~\cite{HuWLNFG21} and pNCNN~\cite{eldesokey2020uncertainty} on fewer points.}
	\vspace{-4mm}
	\label{fig:generation_lines}
\end{figure}

The sparse depth map of the KITTI dataset is obtained by a 64-line Velodyne LiDAR. However, in many practical applications, only 32-line or 16-line LiDAR will be uesed due to cost constraints, which only provide fewer depth points. Therefore, it is crucial to analyze the performance of the method on sparse depth maps with different sparsity levels.
 We trained LRRUs on 64-Line LiDAR depth of KITTI validation set and tested the performance (RMSE[mm]) on fewer lines that is obtained by the method provided by \cite{zhao2021ral}.
Although our results shown in the \figref{fig:generation_lines} gradually decrease with decreasing depth points (from 64-Line to 16-Line), they still outperform competing methods at different sparse patterns,  which shows that our model has better generalization capability on more sparse depth maps.

\subsection{Experiments on NYUv2 Dataset}

To verify the effectiveness of our proposed method in indoor scenes, we conduct comparative experiments on the NYUv2 dataset.
Following existing methods~\cite{MaCK19,tang2019learning,lin2022dynamic}, we train and evaluate our model with the setting of 500 sparse LiDAR samples.
The quantitative evaluation results in \tabref{tab:NYUv2} demonstrate that the four variants of the proposed LRRU achieve the best performance across different parameter regimes.
Specifically, our LRRU-Mini model achieves an impressive results, which employs only 0.3M parameters to obtain the same RMSE as GuideNet~\cite{tang2019learning} using 74M parameters.
We empirically found that if using a UNet-like model with similar amount of parameters (0.3M) to directly regress, the model will fail.
Meanwhile, the LRRU-Base model outperforms state-of-the-art methods with relatively fewer parameters.

\begin{figure}[!t]
	\centering
	\includegraphics[width=0.5\textwidth]{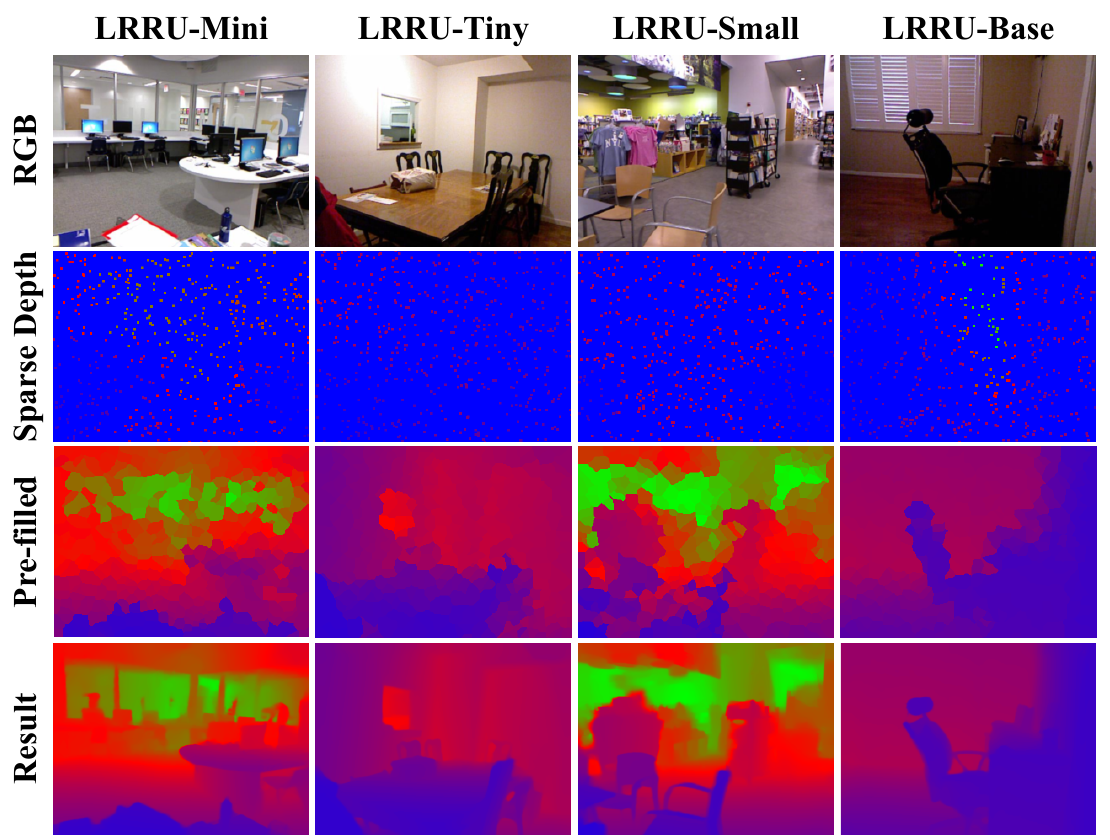}
	\caption{Qualitative results of the proposed LRRUs on the NYUv2 dataset. The sparse depth is dilated to show and ``Pre-filled'' denotes the pre-filled depth map. Since the sparse depth map has few available points, the pre-filled depth is very coarse. However, our LRRUs can update it to satisfactory results.}
	\vspace{-4mm}
	\label{fig:NYUv2_visualized}
\end{figure}

\figref{fig:NYUv2_visualized} shows the qualitative results of the proposed LRRU models. Since the input depth of the NYUv2 dataset is more sparse than the depth of the KITTI dataset, the pre-filled depth map obtained by a simple hand-crafted method~\cite{ku2018defense} looks terrible.
However, LRRUs can rectify the initial depth map by the recurrent update process, which obtains satisfactory results, even in some thin structures.

\begin{table}[!t]
\centering
\footnotesize
\setlength{\tabcolsep}{0.8mm}
    \begin{minipage}{.5\textwidth}
        \centering
        \caption{Quantitative evaluation on the NYUv2 dataset. The best and second-best RMSE and REL results are highlighted in \redfont{red} and \bluefont{blue} colors
        , respectively.}
        \label{tab:NYUv2}
            \begin{tabular}{@{}lrccccc@{}}
            \hline
            Methods  & Params.  & RMSE{[}m{]}         & REL                  & $\delta_{1.25}$     & $\delta_{{1.25}^{2}}$ & $\delta_{{1.25}^{3}}$ \\ \hline \hline
            S2D~\cite{MaCK19}   &  26 M  & 0.230 & 0.044 & 97.1 & 99.4 & 99.8  \\
            CSPN~\cite{cheng2019cspn}  &  17.4 M  & 0.117 & 0.016 & 99.2 & 99.9 & 100.0 \\
            DeepLiDAR~\cite{qiu2019deeplidar} & 48 M & 0.115 & 0.022 & 99.3 & 99.9 & 100.0 \\
            DepthNormal~\cite{xu2019depth} & - M & 0.112 & 0.018 & 99.5 & 99.9 & 100.0 \\
            FCFRNet~\cite{liu2021fcfr} & - M  & 0.106 & 0.015 & 99.5 & 99.9 & 100.0 \\
            ACMNet~\cite{zhao2021adaptive}  & 4.9 M  & 0.105 & 0.015 & 99.4 & 99.9 & 100.0 \\
            GuideNet~\cite{tang2019learning} & 74 M & 0.101 & 0.015 & 99.5 & 99.9 & 100.0 \\
            NLSPN~\cite{park2020non}  & 25.8 M   & 0.092 & \bluefont{0.012} & 99.6 & 99.9 & 100.0 \\ 
            RigNet~\cite{yan2021rignet} & - M   & \redfont{0.090} & \bluefont{0.012} & 99.6 & 99.9 & 100.0 \\ 
            DySPN~\cite{lin2022dynamic}  & $\sim$26 M   & \redfont{0.090} & \bluefont{0.012} & 99.6 & 99.9 & 100.0 \\ 
            \hline
            LRRU-Mini & {0.3 M} & \multicolumn{1}{c}{0.101} & \multicolumn{1}{c}{0.013} & \multicolumn{1}{c}{99.4} & \multicolumn{1}{c}{99.9} & \multicolumn{1}{c}{100.0} \\ 
            LRRU-Tiny  & 1.3 M & \multicolumn{1}{c}{0.096} & \multicolumn{1}{c}{0.012} & \multicolumn{1}{c}{99.5} & \multicolumn{1}{c}{99.9} & \multicolumn{1}{c}{100.0} \\
            LRRU-Small & 5 M  & \multicolumn{1}{c}{0.093} & \multicolumn{1}{c}{0.012} & \multicolumn{1}{c}{99.5} & \multicolumn{1}{c}{99.9} & \multicolumn{1}{c}{100.0} \\
            LRRU-Base & 21 M & \multicolumn{1}{c}{\bluefont{0.091}} & \multicolumn{1}{c}{\redfont{0.011}} & \multicolumn{1}{c}{{99.6}} & \multicolumn{1}{c}{{99.9}} & \multicolumn{1}{c}{{100.0}} \\
            \hline
            \end{tabular}
    \end{minipage}%
\vspace{2mm}

        \begin{minipage}{.5\textwidth}
        \centering
        \caption{Quantitative evaluation on the KITTI validation dataset \\ \textbf{under the depth-only setting}.}
        \label{tab:depth-only}
            \begin{tabular}{@{}lrcc@{}}
            \hline
            Methods &
              Params.~~~ &
              \begin{tabular}[c]{@{}c@{}}~~RMSE {[}mm{]}~~~~\end{tabular} &
              \begin{tabular}[c]{@{}c@{}}~~MAE {[}mm{]}~~\end{tabular} \\ \hline \hline
            S2D (Depth-only)~\cite{MaCK19}~~~  & 26 M~~~  & 985.1      & 286.5          \\          
            CUNet~\cite{wang2022ral}~  & 35 M~~~    & 958.8 & 245.3  \\ \hline
            LRRU-Mini  & \textbf{0.3 M~~~}  & 987.7 & 251.8   \\
            LRRU-Tiny  & 1.3 M~~~  & 958.2 & 242.4  \\
            LRRU-Small~ & 5 M~~~  & \textbf{950.9} & 237.1   \\
            LRRU-Base  & 21 M~~~  & 957.4 & \textbf{235.9}   \\ \hline
            \end{tabular}
    \end{minipage}
\vspace{-6mm}
\end{table}

\subsection{Extension to Depth-only Case}

As a general framework for depth completion, our approach can be extended to scenarios where only sparse depth maps are available.
In the depth-only setting, we only extract the cross-guided feature from the sparse input depth map.
We report the performance of our method on the KITTI validation dataset in \tabref{tab:depth-only}.
The experimental results show that the method performance drops significantly compared to using RGB image input.
We consider that this is because if only the sparse depth map is input, the information provided is very limited.
It is difficult for the depth completion methods to recover the structural details of the depth map.
However, it is worth noting that our method still achieves better results compared with S2D (Depth-only)~\cite{MaCK19} and CUNet\cite{wang2022ral}, where CUNet is specifically designed for the depth-only case and achieves the best performance at present.
Meanwhile, we notice that the results of LRRU-Base decrease due to over-fitting.

\section{Conclusion}\label{sec::conclusion}
In this paper, we have proposed Long-short Range Recurrent Updating (LRRU) network, a novel lightweight deep network framework for depth completion. 
Based on the proposed target-dependent update unit and long-short range recurrent strategy, our iterative update process is content-adaptive and highly flexible.
Compared with conventional direct-regression approaches, our method achieves superior performance with fewer parameters and inference time.
Experimental results demonstrate that our methods outperform existing methods on both indoor and outdoor scenes, and achieve SOTA performance across different parameter regimes. In the future, we will explore our method for other dense prediction tasks, such as monocular depth estimation and semantic segmentation.

\section*{Acknowledgments}
This research was supported in part by the National Natural Science Foundation of China (62001394, 62271410), and the Fundamental Research Funds for the Central Universities.

{\small
\bibliographystyle{ieee_fullname}
\bibliography{egbib}
}

\end{document}


\title{LRRU: Long-short Range Recurrent Updating Networks for Depth Completion \\ -- Supplementary Materials --}

\def\SP{~~}

\author{Yufei Wang$^{1}$,
\SP
Yuchao Dai$^{1}$\thanks{Corresponding author.},
\SP
Ge Zhang$^{1}$,
\SP
Bo Li$^{1}$,
\SP
Qi Liu$^{1}$,
\SP
Tao Gao$^{2}$
\\[0.1325cm]
$ ^1$Northwestern Polytechnical University and Shaanxi Key Laboratory of\\  Information Acquisition and Processing
\SP $^2$Chang'an University 
}

\maketitle

\begin{abstract}
In this supplementary material, we provide further details in evaluation metrics, quantitative and qualitative analysis of our iterative update process, and qualitative comparisons on the NYUv2 dataset. 
In addition, we adopt the proposed LRRU method in the basic framework of existing SPN-based methods to further demonstrate the advantages of our method.
\end{abstract}

\section{Evaluation metrics.} 

Following exiting depth completion methods~\cite{MaCK19,lin2022dynamic, cheng2019cspn,cheng2020cspn++,xu2020deformable,park2020non,HuWLNFG21,lin2022dynamic,liu2022graphcspn}, we employ the Root Mean Squared Error ({RMSE}[$\mathrm{mm}$]), Mean Absolute Error ({MAE}[$\mathrm{mm}$]), Root Mean Squared Error of the Inverse depth ({iRMSE}[1/ $\mathrm{km}$]), Mean Absolute Error of the Inverse depth ({iMAE}[1/$\mathrm{km}$]), mean absolute relative error ({REL}), and percentage of pixels satisfying \textbf{$\delta_{\tau}$} for quantitative evaluation. 
Eq.~\eqref{equ:metrics} shows the detailed definitions, where $d^{gt}$ denotes the ground truth depth map, $d^{pred}$ denotes the predicted dense depth map, and $\mathcal{V}$ is the set of available points in the ground truth.

\section{Analysis of Our Iterative Update Process.}

In this paper, we present the Long-short Range Recurrent Updating Network (LRRU), a novel lightweight deep network framework for depth completion. Given a sparse input depth map, we first fill it to an initial dense depth map using a simple non-learning method~\cite{ku2018defense}.
Then, according to the long-short range recurrent strategy, our method iteratively updates the initial depth map through the Target-Dependent Update Unit (TDU) to obtain the final accurate depth maps.
Our update process is target-adaptive and highly flexible, and the proposed method only requires 4 iterations to achieve satisfactory results.
In this section, we perform both qualitative and quantitative analysis on the result of the depth map during the iterative update process.
The experiments are conducted on the KITTI validation dataset~\cite{Geiger2012CVPR}.

\begin{equation}[!t]
\label{equ:metrics}
\begin{aligned}
&\operatorname{RMSE} [\mathrm{mm}]: \sqrt{\frac{1}{|\mathcal{V}|} \sum_{v \in \mathcal{V}}\left|d_{v}^{g t}-d_{v}^{pred}\right|^{2}}, \\
&\operatorname{MAE} [\mathrm{mm}]:  \frac{1}{|\mathcal{V}|} \sum_{v \in \mathcal{V}}\left|d_{v}^{g t}-d_{v}^{\text {pred }}\right|, \\
&\operatorname{iRMSE} [1 / \mathrm{km}]: \sqrt{\frac{1}{|\mathcal{V}|} \sum_{v \in \mathcal{V}}\left|1 / d_{v}^{g t}-1 / d_{v}^{\text {pred }}\right|^{2}}, \\
& \operatorname{iMAE} [1 / \mathrm{km}]: \frac{1}{|\mathcal{V}|} \sum_{v \in \mathcal{V}}\left|1 / d_{v}^{g t}-1 / d_{v}^{\text {pred }}\right|, \\
& \text { REL }: \frac{1}{|\mathcal{V}|} \sum_{v \in \mathcal{V}}\left|\left(d_v^{g t}-d_v^{p r e d}\right) / d_v^{g t}\right|, \\
&\delta_\tau [\%]:  \max \left(\frac{d_v^{g t}}{d_v^{p r e d}}, \frac{d_v^{p r e d}}{d_v^{g t}}\right)<\tau.
\end{aligned}
\end{equation}

\subsection{Quantitative Analysis}

\begin{figure}[!t]
\centering
\begin{minipage}[b]{.45\textwidth}
\centerline{\includegraphics[width=\linewidth]{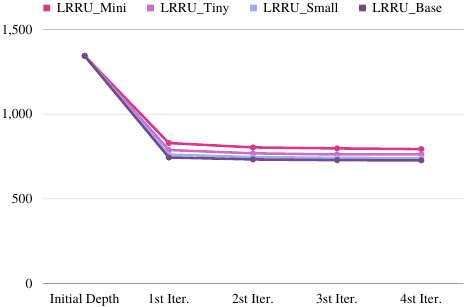}}
\caption{The RMSE [mm] change of the depth map during the iterative update process.}\label{iteration-a}
\vspace{2mm}
\end{minipage}\qquad
\begin{minipage}[b]{.45\textwidth}
\centerline{\includegraphics[width=\linewidth]{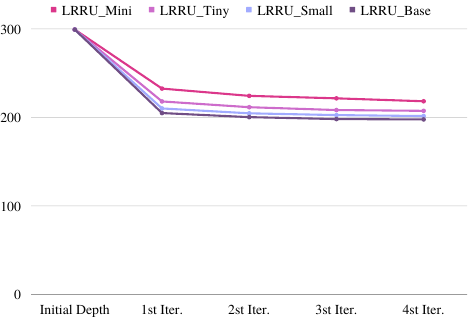}}
\caption{The MAE [mm] change of the depth map during the iterative update process.}\label{iteration-b}
\end{minipage}
\vspace{-4mm}
\end{figure}

In our proposed method, the accuracy of the initial dense depth map is not strictly limited. We obtain the initial depth map by a classical handcrafted approach~\cite{ku2018defense}. \figref{iteration-a} and \figref{iteration-b} show the quantitative results of the depth map during the update process, which include RMSE and MAE.
The experimental results demonstrate that the initial depth map is coarse, and the RMSE and MAE are 1344 mm and 299.3 mm, respectively.
However, when we update the initial depth once through our Target-Dependent Update Unit (TDU), the results of the depth map are greatly enhanced.
Then, as the depth map is updated iteratively, its performance is continuously improved.

\subsection{Qualitative Analysis}

\begin{figure*}
	\centering
	\includegraphics[width=0.95\textwidth]{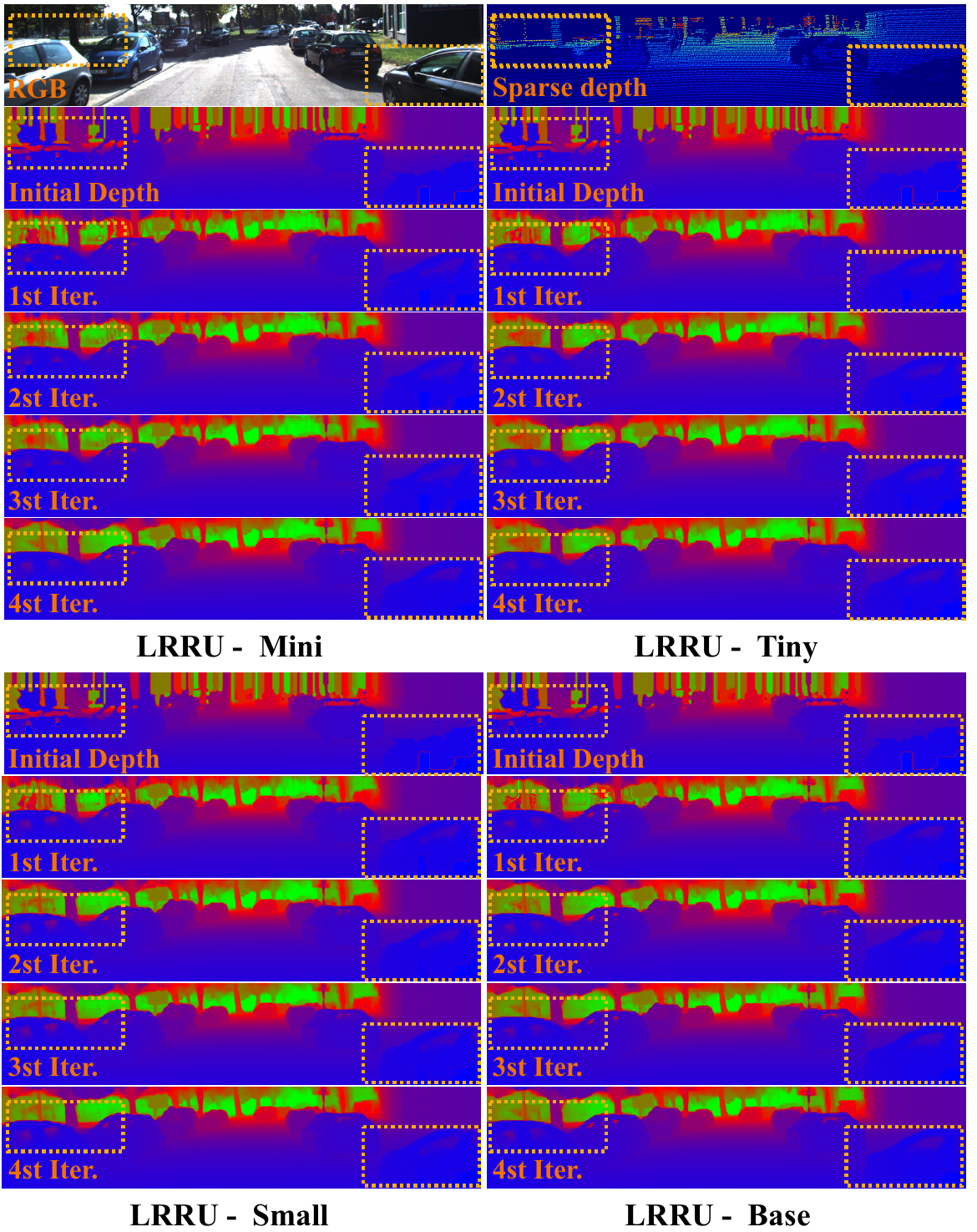}
	\caption{\textbf{Typical examples to demonstrate the qualitative result of the depth map during our iterative update process.} Note that the sparse depth has been dilated for better visualization.}
	\label{fig:iteration_example}
\end{figure*}

\figref{fig:iteration_example} shows some typical examples that demonstrate the qualitative results of the depth map during our iterative update process.
As described in this paper, our proposed method employs large-to-small kernel scopes to capture long-to-short range dependencies to update the depth map.
When we update the initial coarse depth map by a TDU with a large kernel scope, the structural details of the depth map are greatly improved.
However, the object boundaries of the depth image are still blurred.
Then, we iteratively update the depth map by TDUs with smaller kernel scopes, and the depth map continues to become more refined.
Meanwhile, the experimental results demonstrate that the larger the network size of our proposed method, the finer the depth map at the same stage.

\section{Qualitative Comparisons on NYUv2.}

\begin{figure*}
	\centering
	\includegraphics[width=0.95\textwidth]{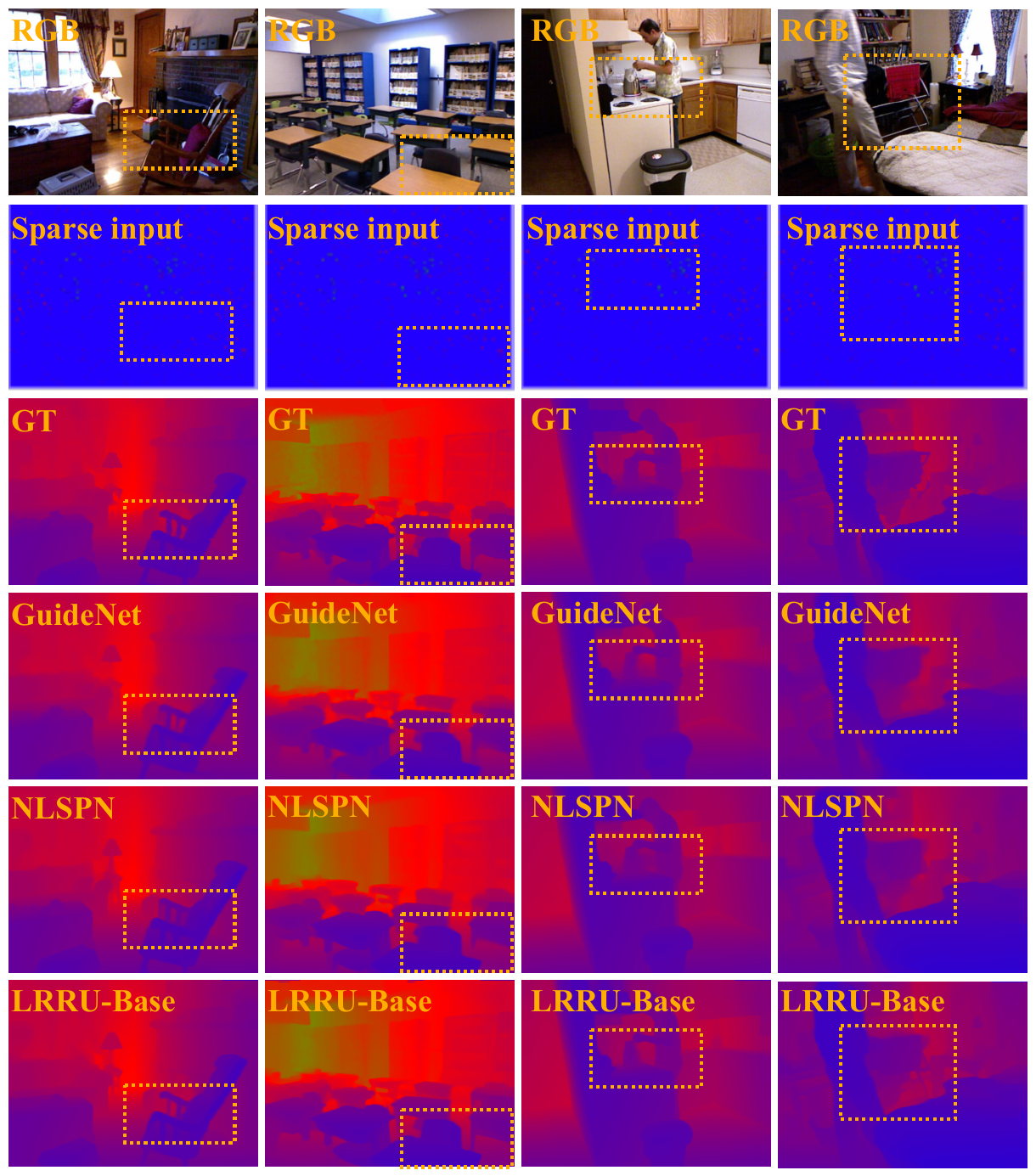}
	\caption{\textbf{Qualitative comparison with SOTA methods on the NYUv2 dataset.} From top to bottom are RGB images, sparse input depth maps, ground truth depth maps, dense depth maps predicted by GuideNet~\cite{tang2019learning}, NLSPN~\cite{park2020non} and our LRRU-Base model, respectively. Note that the sparse input has been dilated for better visualization.}
	\vspace{12mm}
	\label{fig:NYUv2}
\end{figure*}

In \figref{fig:NYUv2}, we present some depth completion results predicted by our method and other state-of-the-art (SOTA) methods, such as GuideNet~\cite{tang2019learning} and NLSPN~\cite{park2020non}, on the indoor NYUv2 depth dataset.
Following the previous works~\cite{MaCK19,lin2022dynamic, cheng2019cspn,cheng2020cspn++,xu2020deformable,park2020non,HuWLNFG21,lin2022dynamic,liu2022graphcspn}, the sparse depth maps are obtained by randomly sampling 500 depth pixels from the dense depth maps and fed into the network with corresponding RGB images.
The experimental results demonstrate that GuideNet generates blurry depth maps, as it is a direct regression method.
Compared to GuideNet, the depth maps predicted by NLSPN and our method substantially improve accuracy thanks to the iterative update scheme.
However, NLSPN suffers from over-smoothing problems due to massive iterations, especially on tiny or thin structures.
In contrast, our method preserves tiny structures and depth boundaries well by using the proposed Target-Dependent Update Unit (TDU) and long-short range recurrent strategy.

\section{LRRU using SPN framework (LRRU-SPN)}

\begin{figure*}
	\centering
	\includegraphics[width=0.87\textwidth]{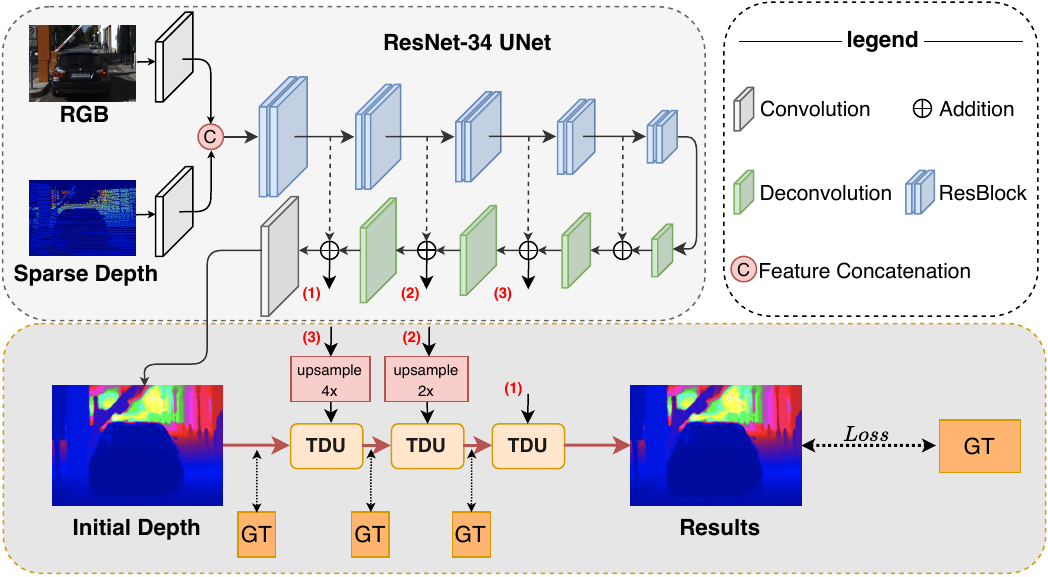}
	\caption{\textbf{The network architecture of our LRRU-SPN.} For a fair comparison, we employ the same network structure as the SPN-based methods. We replace the spatial propagation networks with our \textbf{Target-Dependent Update Unit (TDU)} and iteratively employ it to update the output of the encoder-decoder network 3 times according to the long-short range recurrent strategy.}
	\label{fig:LRRU-SPN}
\end{figure*}

For a direct comparison with existing SPN-based methods~\cite{cheng2019cspn,cheng2020cspn++,xu2020deformable,park2020non,HuWLNFG21,lin2022dynamic,liu2022graphcspn}, we adopt the proposed LRRU method in the basic framework of the SPN-based methods.
We name our method LRRU-SPN with a little name misuse.
As shown in \figref{fig:LRRU-SPN}, LRRU-SPN uses the ResNet-34~\cite{he_ResNet_CVPR_2016} as the backbone in the same way as these SPN-based approaches.
In addition, LRRU-SPN replaces the spatial propagation networks with our proposed target-dependent update unit, and iteratively employ it to update the output of the encoder-decoder network according to our long-short range recurrent strategy.

Conventional SPNs (\eg CSPN~\cite{cheng2019cspn} and NLSPN~\cite{park2020non}) employ fixed kernel weights and scopes, which require massive iterations to obtain the long-range dependencies and decent results.
By contrast, LRRU-SPN adaptively estimates the kernel weights and dynamically obtains long-to-short range dependencies by adjusting the kernel scope from large to small, which is more flexible.
In this comparison experiment, we only need three iterations, where the TDUs of the update process are guided by the cross-guided features of $1/4-$scale, $1/2-$scale, and full-scale in turn.
The results in \tabref{tab:spn} show that LRRU-SPN outperforms existing SPN-based methods and has faster inference time.

Although the process of dynamically predicting the  kernel parameters will consume extra time, our method is not constrained by the number of iterations.
The results on KITTI validation set show that our method using only one iteration (RMSE:741.8\textbf{/}time:0.07s) already outperforms NLSPN (RMSE:771.8/time:0.16s).
Therefore, we consider our method to be more efficient.

\begin{table}[htbp]
    \small
    \setlength{\tabcolsep}{0.8mm}
   \begin{minipage}{.5\textwidth}
        \centering
        \caption{Quantitative evaluation with SPN-based methods on KTTTI test dataset. The time$^{*}$ are from the KITTI leaderboard due to the lack of open-source code.}
        \label{tab:spn}
        \begin{tabular}{lclcc}
        \hline
        SPNs & Iterations & Time{[}s{]} & RMSE{[}mm{]} & MAE{[}mm{]} \\ \hline \hline
        CSPN~\cite{cheng2019cspn}        & 12        &  ~0.20s            & 1019.64       & 279.46       \\
        DSPN~\cite{xu2020deformable}        & 12        &   ~0.34s$^{*}$       & 766.74        & 220.36       \\
        NLSPN~\cite{park2020non}       & 18        &   ~0.13s           & 741.68        & 199.59       \\
        CSPN++~\cite{cheng2020cspn++}& 12        &      ~0.20s$^{*}$        & 743.69        & 209.28       \\
        DySPN~\cite{lin2022dynamic}       & 6         &    ~0.16s$^{*}$         & 709.12        & \textbf{192.71}       \\
        GraphCSPN~\cite{liu2022graphcspn}   & 3         &    ~0.12s$^{*}$          & 736.24        & 193.25       \\ \hline
        LRRU-SPN    & \textbf{3}         &     \textbf{~0.10s}         &      \textbf{707.11}          &    202.08          \\ \hline
        \end{tabular}
        
    \end{minipage}
\end{table}

\newpage

{\small
\bibliographystyle{ieee_fullname}
\bibliography{egbib}
}